\documentclass[runningheads]{llncs}
\usepackage{svg}
\usepackage{amsmath} 
\usepackage{graphicx}
\begin{document}

\title{Efficient Large Scale Medical Image Dataset Preparation for Machine Learning Applications}
\titlerunning{Efficient Large Scale Medical Image Dataset Preparation}
%
\author{
Stefan Denner\inst{1, 2} \and
Jonas Scherer\inst{1} \and
Klaus Kades\inst{1} \and
Dimitrios Bounias\inst{1,2} \and
Philipp Schader\inst{1} \and
Lisa Kausch\inst{1} \and
Markus Bujotzek\inst{1,2} \and
Andreas Michael Bucher\inst{5} \and
Tobias Penzkofer\inst{6} \and
Klaus Maier-Hein\inst{1, 3, 4}
}

\authorrunning{S. Denner et al.}
%
\institute{
Division of Medical Image Computing, German Cancer Research Center (DKFZ), Heidelberg, Germany \and
Medical Faculty Heidelberg, University of Heidelberg, Heidelberg, Germany \and
Pattern Analysis and Learning Group, Department of Radiation Oncology, Heidelberg University Hospital, Heidelberg, Germany \and
National Center for Tumor Diseases (NCT) Heidelberg, Germany \and
Goethe University Frankfurt, University Hospital, Institute for Diagnostic and Interventional Radiology, Frankfurt am Main, Germany \and
Department of Radiology, Charité - Universitätsmedizin Berlin, Berlin, Germany
}
\maketitle              
\setcounter{footnote}{0}

\begin{abstract}
In the rapidly evolving field of medical imaging, machine learning algorithms have become indispensable for enhancing diagnostic accuracy. 
However, the effectiveness of these algorithms is contingent upon the availability and organization of high-quality medical imaging datasets. 
Traditional Digital Imaging and Communications in Medicine (DICOM) data management systems are inadequate for handling the scale and complexity of data required to be facilitated in machine learning algorithms. 
This paper introduces an innovative data curation tool, developed as part of the Kaapana\footnote{\url{https://github.com/kaapana/kaapana}} open-source toolkit, aimed at streamlining the organization, management, and processing of large-scale medical imaging datasets. 
The tool is specifically tailored to meet the needs of radiologists and machine learning researchers. 
It incorporates advanced search, auto-annotation and efficient tagging functionalities for improved data curation.
Additionally, the tool facilitates quality control and review, enabling researchers to validate image and segmentation quality in large datasets. 
It also plays a critical role in uncovering potential biases in datasets by aggregating and visualizing metadata, which is essential for developing robust machine learning models. 
Furthermore, Kaapana is integrated within the Radiological Cooperative Network (RACOON), a pioneering initiative aimed at creating a comprehensive national infrastructure for the aggregation, transmission, and consolidation of radiological data across all university clinics throughout Germany. 

A supplementary video showcasing the tool's functionalities can be accessed at \url{https://bit.ly/MICCAI-DEMI2023}.

\end{abstract}
%
%
%


\begin{figure}[h!]
  \centering
  \includegraphics[width=\textwidth]{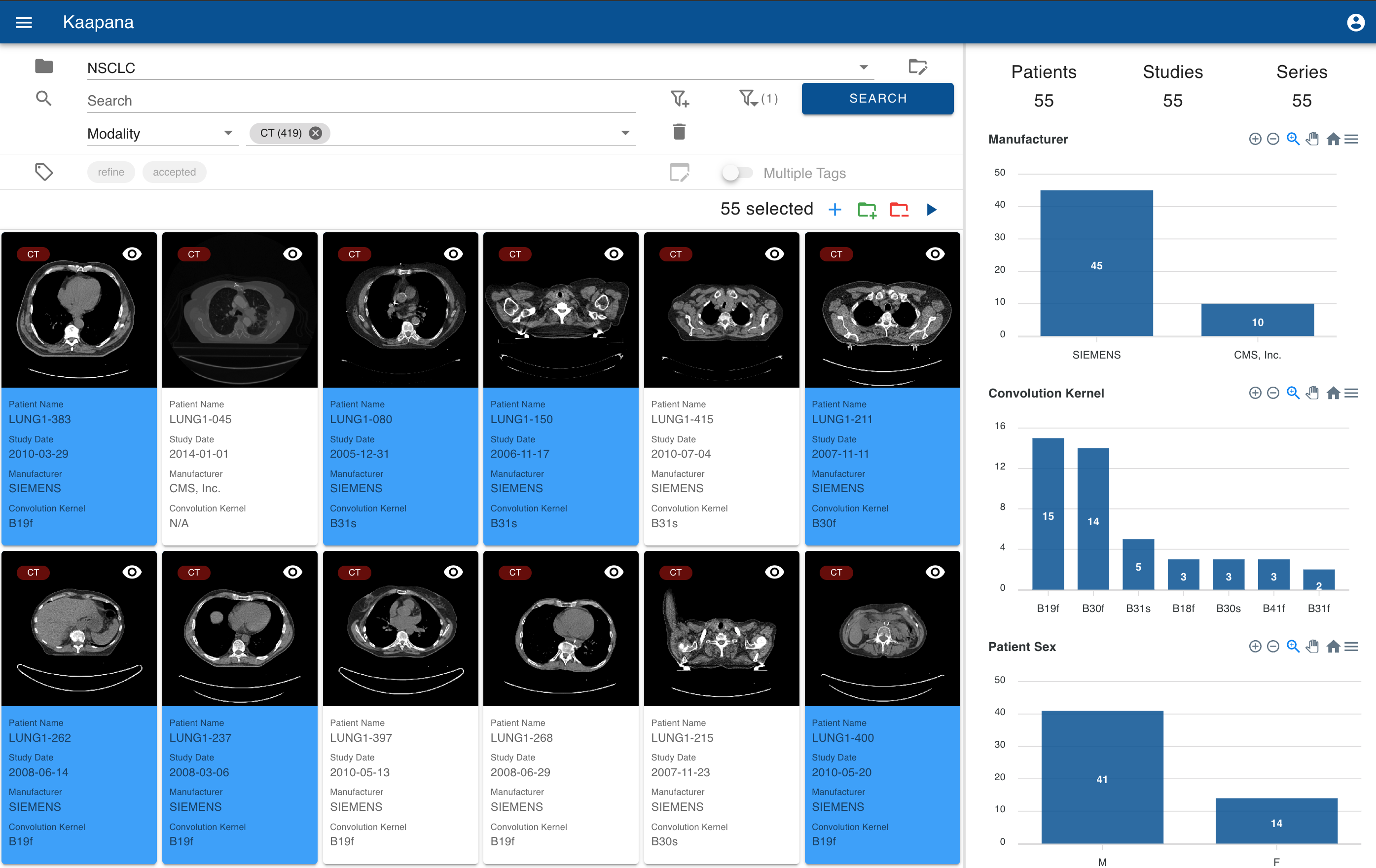}
  \caption{Screenshot of the curation tool integrated into Kaapana. The gallery view displays series thumbnails accompanied by customizable metadata, providing a comprehensive visual overview. The sidebar showcases the metadata of the current selection, enabling swift detection of potential biases based on the DICOM metadata. This layout illustrates the tool's user-friendly interface and its capabilities in efficient data curation and bias detection.}
  \label{fig:screenshot_with_metadata}
\end{figure}

\section{Introduction}
In recent years, the development and application of machine learning algorithms in medical imaging have emerged as an instrumental component in advancing healthcare and diagnostic accuracy \cite{advancementsInMI}. 
This advancement, however, depends heavily on the availability and organization of high-quality medical imaging datasets \cite{performanceImprovesWithMoreData,shortcutCovid19}.
The Digital Imaging and Communications in Medicine (DICOM) standard, commonly adopted for storing medical images, encapsulates both image data and vital metadata, including image modality, acquisition device manufacturer, and patient information like age and gender \cite{dicom,dicom2}. 
This metadata holds considerable value in the development of robust medical imaging machine learning algorithms \cite{datasetBias}.
Traditional DICOM data management systems, while effective for individual scans or patients, struggle to efficiently handle the scale and complexity of the data needed to be facilitated in machine learning algorithms \cite{OHIFViewer}. 
The demand for superior data curation tools is crucial for advancing the field of medical imaging \cite{summersPreparing,magudia2021trials}. 
Despite recent progress in medical imaging data curation, existing solutions exhibit certain limitations. Some tools, while useful for data curation, are either proprietary, ill-equipped to handle large-scale medical datasets, or fail to fully exploit the benefits of DICOM headers \cite{curationTools}.

Additionally, while there are automated approaches to enhance the data curation process, they are not conveniently integrated into a user-friendly tool \cite{automatedCuration,wasserthal2022totalsegmentator,tang2021body}. 

In response to these challenges, we have developed an innovative data curation tool as part of the Kaapana open-source toolkit \cite{scherer2020joint,kaapana}. 
Kaapana is designed for advanced medical data analysis, especially in radiological and radiotherapeutic imaging, facilitating AI-driven workflows and federated learning approaches. 
By enabling on-site data processing and ensuring seamless integration with clinical IT infrastructures, it aims to address challenges in multi-center data acquisition and offers tools for standardized data processing workflows, distributed method development, and large-scale multi-center studies.
Building up on Kaapana, our tool is designed to streamline the organization, management, and processing of large-scale medical imaging datasets, catering specifically to the needs of radiologists and machine learning researchers. 

Our contribution is threefold: Our data management tool facilitates (1) efficient data curation by advanced search, auto-annotation and tagging, (2) quality control and review and (3) dataset bias detection by metadata visualization.

Kaapana is a constituent of the Radiological Cooperative Network (RACOON), an initiative to establish a nationwide infrastructure for collecting, transferring, and pooling radiological data across all German university clinics. Integrating our tool in Kaapana paves the way for its imminent deployment across all German university clinics, facilitating clinical validation.

\section{Methodology}
The essence of our methodology is to extend the capabilities of Kaapana, an open-source toolkit designed for medical data analysis and platform provisioning, by incorporating a comprehensive tool for managing, curating, and processing large-scale medical imaging datasets.

\begin{figure}[h!]
  \centering
  \includegraphics[width=\textwidth]{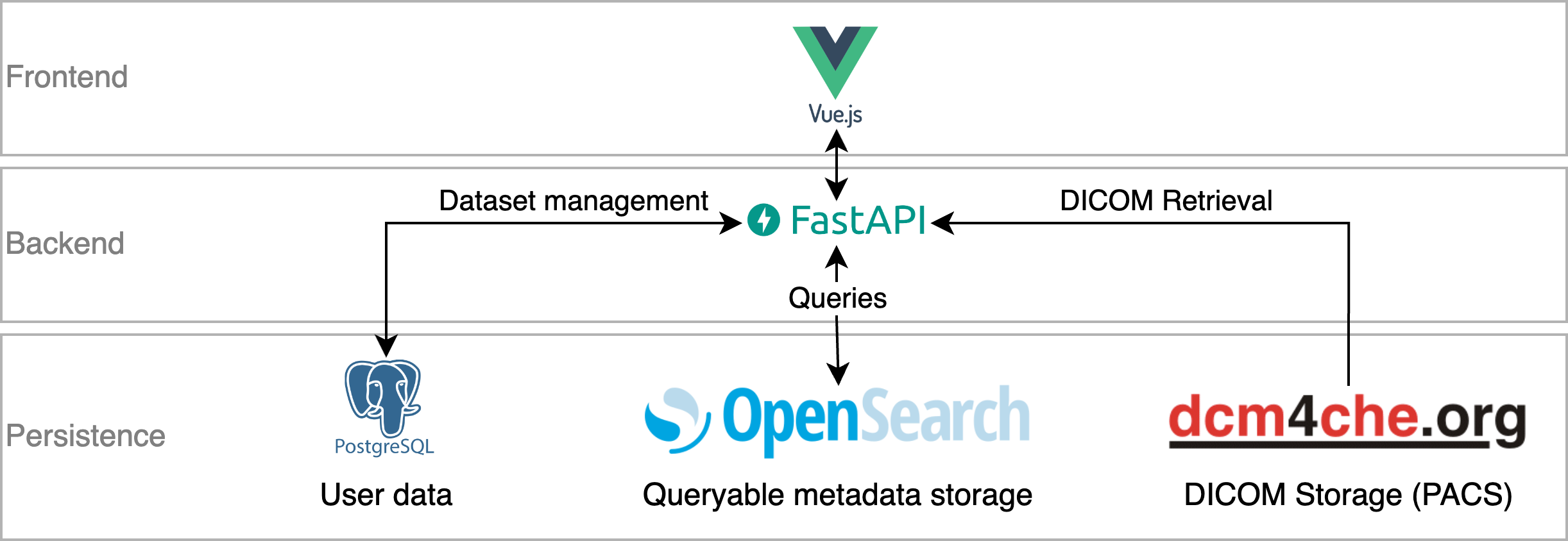}
  \caption{Illustrating the technical infrastructure. Highlighting the frontend powered by Vue.js, the backend using FastAPI, and the three-fold data persistence layer consisting of dcm4chee, OpenSearch, and PostgreSQL. The arrows visualize the communication between the components.}
  \label{fig:architecture}
\end{figure}

\subsection{Technical Infrastructure}

Our system benefits from Kaapana's robust technical infrastructure. 
Vue.js, a versatile JavaScript framework, powers the frontend, ensuring user-friendly, dynamic, and responsive web interfaces. 
FastAPI, a high-performance web framework, forms the backbone of the backend, enabling efficient communication with the frontend.
\newline
The persistence layer is three-fold, each serving a unique purpose. 
The dcm4chee Picture Archiving and Communication System (PACS) stores the original DICOM images, safeguarding their integrity and availability. 
For efficient management of large datasets, the DICOM Header is converted to JSON and stored in OpenSearch, a powerful open-source search engine known for its quick querying abilities. 
PostgreSQL, an open-source object-relational database system, forms the mapping layer, establishing connections between data and respective datasets, hence facilitating effective categorization and retrieval.
The full utilized technical infrastructure is visualized in Fig. \ref{fig:architecture}.

While our focus has primarily been on DICOM data, our solution also demonstrates flexibility in accommodating other formats. Kaapana is capable of transforming images in the Neuroimaging Informatics Technology Initiative (NIfTI) data format into DICOMs. These transformed images can then be curated. It's important to note, however, that metadata extraction is not possible from the NIfTI format; only the image data is preserved in the transformation. Nevertheless, this flexibility in data handling further extends the applicability of our tool in a variety of medical imaging contexts.

\subsection{Graphical User Interface}

\begin{figure}[ht]
    \centering
    \includegraphics[width=\textwidth]{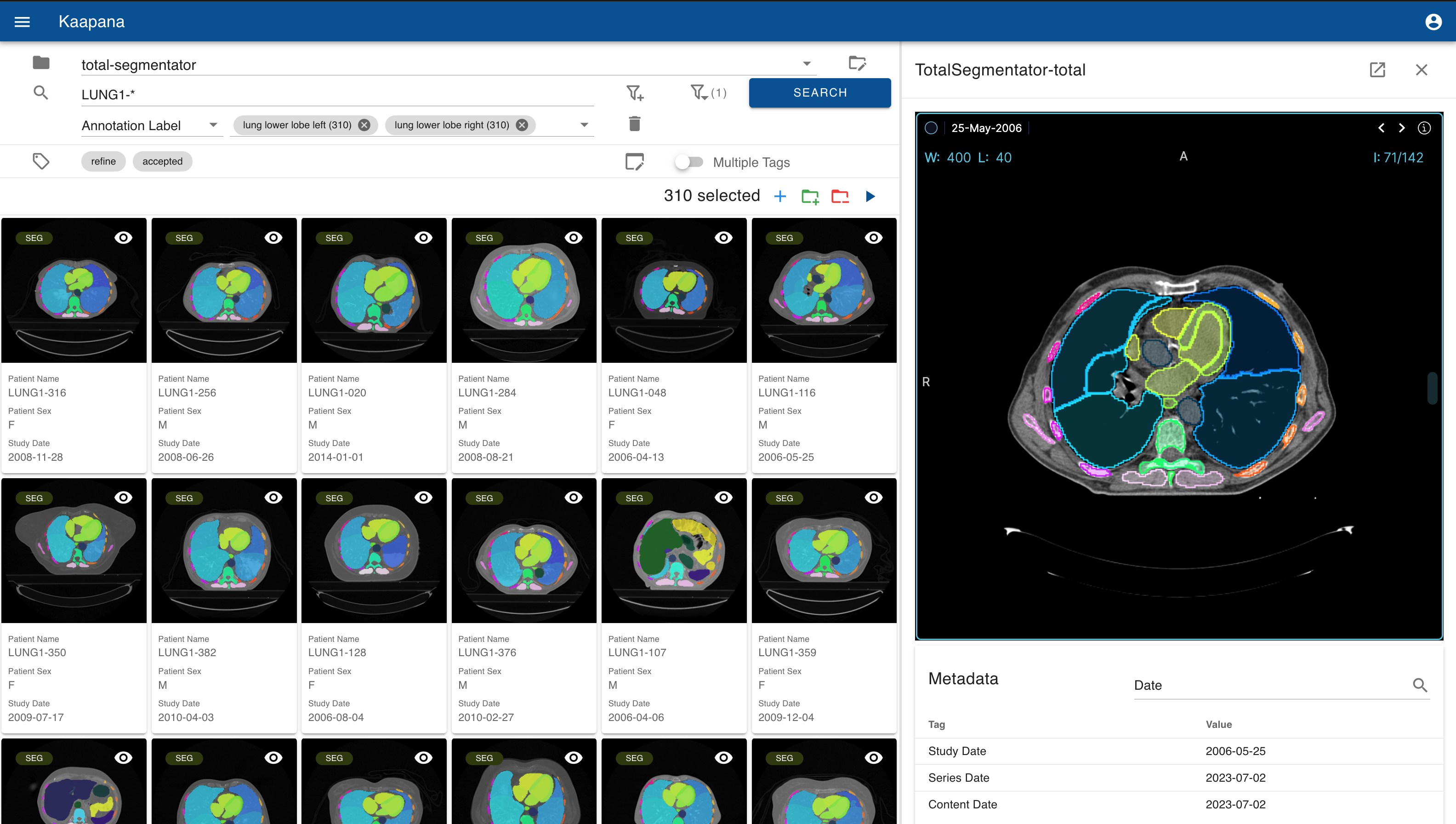}
    \caption{Filtering for series containing lower lung lobes. The gallery view presents thumbnails with superimposed segmentations, while one selected series is opened in the sidebar for interactive 3D volume visualization with segmentations.}
    \label{fig:screenshot}
\end{figure}

The graphical user interface in seamlessly integrated into Kaapana's Vue.js frontend. 
Throughout the development process, which was conducted in close collaboration with radiologists, it was highlighted that varying use cases necessitate distinct user interface requirements. 
Consequently, the user interface has been designed to be highly adaptable, offering an array of customizable settings to cater to diverse needs.
Overall, the user interface consists of a three-part layout visualized in Figure \ref{fig:screenshot}.
\subsubsection{Search.}
A sophisticated full-text search function, supporting wildcard search and free-text filtering, assists users in efficiently locating specific items based on image metadata. 
Additionally, it provides autocomplete functionality, streamlining the search process.

\subsubsection{Gallery View.}
The gallery view provides a visual display of DICOM series, presenting them in a thumbnail format along with customizable metadata. 
The thumbnail creation is in compliance with the DICOM standard \cite{dicom,dicom2}, which accommodates a broad spectrum of image modalities, including but not limited to, Structured Reports (SR), CT, or Magnetic Resonance Imaging (MRI). 
Given the current interest in segmentation algorithms within the medical imaging community \cite{maier2018rankings}, our tool automatically generates thumbnails for DICOM-SEGs or RTStructs that illustrates the segmentation superimposed on the original image. 
The gallery view also includes a multi-selection feature that facilitates bulk operations (see Fig. \ref{fig:screenshot_with_metadata}). 

\subsubsection{Sidebar.}
The sidebar serves a dual function - as a metadata dashboard and a detail view. The configurable metadata dashboard aggregates and displays comprehensive metadata distributions based on the current selection in the gallery view. These metadata distributions are interactive, allowing for selection and zooming for detailed examination. They can also be downloaded as charts or CSV files, providing flexibility in data analysis and sharing.

In the detail view mode, activated upon series selection, it showcases an interactive 3D visualization of the chosen DICOM series using the integrated (adjusted) OHIF Viewer \cite{OHIFViewer} next to a searchable table with the series' metadata, including the DICOM Headers.

\subsection{Machine Learning Integration}
Kaapana is capable of executing state-of-the-art machine learning algorithms robustly. 
It is already equipped with a robust body part regression algorithm, allowing automatic assignment of which body part is covered in a given CT image \cite{schuhegger2021body,tang2021body}.
Since one of Kaapana's major strengths is the easy extendability, we integrated TotalSegmentator \cite{wasserthal2022totalsegmentator}
to even further extend the automatic data curation capabilities.
TotalSegmentator is based on nnUNet \cite{isensee2021nnu}, an automatically adapting semantic segmentation method, which allows segmenting 104 anatomical structures (27 organs, 59 bones, 10 muscles, 8 vessels) from CT images. 
This integration significantly enhances the automatic annotation capabilities.
Furthermore, by this, users can filter for those body parts or anatomical structures and even further speed up their curation process.

\subsection{Data Management and Workflow Execution}

Our tool incorporates robust data management and workflow execution capabilities. Users can perform various actions on multiple selected series simultaneously, such as adding or removing series from a dataset and initiating workflows. 
An intuitive tagging system, with shortcut and autocomplete support, streamlines data annotation and categorization.

\section{Results}
Our data curation tool, integrated into Kaapana, provides a comprehensive and intuitive interface for managing, organizing, and processing extensive medical imaging datasets, thereby contributing significantly to efficient dataset curation for machine learning algorithms. Here, we highlight potential applications of our tool through a series of illustrative examples:

\subsection{Dataset Management, Auto-Annotation and Tagging}
Radiologists frequently handle vast collections of medical images, encompassing multiple patients, studies, and imaging modalities \cite{magudia2021trials}. A common scenario involves a radiologist tasked with organizing thousands of CT and MRI scans acquired over several years for a large-scale study. Concurrently, in large-scale medical imaging studies curating and annotating an extensive collection of CT scans presents a formidable challenge. This requires a meticulous analysis of thousands of scans for visible disease symptoms, a process that is both labor-intensive and time-consuming.

Our tool offers a solution to these challenges with its gallery-style view, multi-select functionality, and advanced search features. Radiologists can swiftly sift through images, categorizing them into different datasets based on various attributes, such as patient demographics, study type, or imaging modality. The tool's advanced search functionality enables efficient image curation by allowing filters for DICOM metadata or algorithm outcomes, such as body part or anatomical structure.

These machine-assisted annotations provide an initial dataset that radiologists can validate and refine, significantly reducing the manual labor required and streamlining the annotation process. Furthermore, the gallery view, coupled with tagging functionality, enhances the organization of the curated and annotated dataset. This integrated approach to data organization, management, and annotation significantly alleviates the burden on radiologists and accelerates the preparation of data for machine learning applications.

\begin{figure}[h!]
    \centering
    \includegraphics[width=\textwidth]{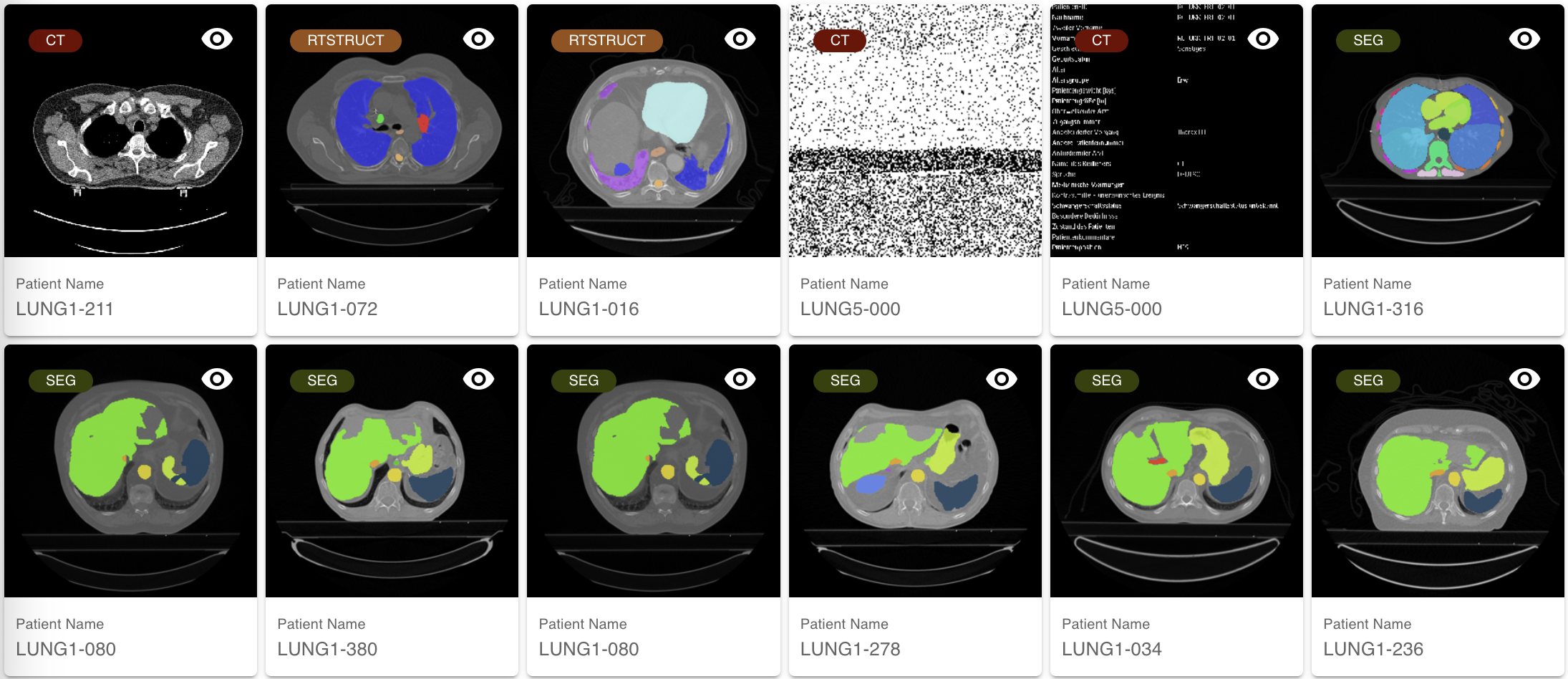}
    \caption{
    Showcasing the gallery view's ability to handle various DICOMs and visually inspecting problematic series, such as the noisy series (top row, fourth from left) and the adjacent patient report. The radiologist can then exclude those problematic series.
    The lower row emphasizes the tool's capacity to quickly spot low-quality segmentations of a 3D CT image.
    }
    \label{fig:segquality}
\end{figure}

\subsection{Quality Control and Review}
Our tool is particularly beneficial in scenarios where researchers need to validate the quality of images and segmentations in large medical imaging datasets, such as those obtained from multi-center studies. The tool's gallery and detail views can be effectively utilized to swiftly pinpoint images with poor quality or erroneous segmentations.
An illustration of this capability is evident in the lower row of Fig. \ref{fig:segquality}, where a multi-organ segmentation algorithm was applied to CT images but yielded subpar results. While 2D thumbnails may not always suffice for quality control of 3D segmentation algorithms, they can significantly expedite the quality control process in certain cases.

For instances where thumbnails fall short, the detail view allows researchers to navigate through the 3D volumes for a more comprehensive quality assurance. 
Moreover, as Magudia et al.\cite{magudia2021trials} highlight, quality control for DICOM Headers is particularly crucial in multi-center studies due to data heterogeneity. 
Our tool caters to this need by displaying and allowing filtering of metadata.

\subsection{Uncovering Potential Bias in Datasets}
Dataset biases, such as disparities in patient demographics or variations in scanner types and configurations, can profoundly influence the performance of machine learning models \cite{datasetBias}. 
Such biases may lead to models that exhibit excellent performance during training and validation phases but falter in real-world applications due to an over-dependence on biased features. For example, a model predominantly trained on data from a specific scanner may struggle to generalize to images produced by other scanners \cite{glocker2019machine}. Our tool can play a pivotal role in identifying these biases through its metadata dashboard. By aggregating and visualizing the metadata of selected items, researchers can discern patterns or inconsistencies that could signal potential biases. The visualization of the metadata distribution from a subset of the LIDC-IDRI Dataset's CT scans, as shown in Fig. \ref{fig:screenshot_with_metadata}, underscores the tool's ability to detect such biases \cite{lidc}. A machine learning model trained on this dataset might inadvertently learn the skewed distribution of convolution kernels or scanners, which could result in failure on unseen data which does not represent the learned distribution.

By offering early detection of bias, the tool enables researchers to implement corrective strategies, such as data augmentation or bias mitigation techniques. This enhances the generalizability and resilience of the machine learning models developed, ensuring they perform optimally across varied scenarios.



\section{Discussion and Conclusion}
The development of an efficient data curation tool as part of the Kaapana open-source toolkit, as presented in this paper, addresses a critical need in the field of medical imaging. 
The availability and organization of high-quality medical imaging datasets are paramount for the successful application of machine learning algorithms in healthcare. 
The tool's integration with Kaapana provides a robust infrastructure for managing, curating, and processing large-scale medical imaging datasets.

One of the significant contributions of this tool is the streamlined annotation process. 
By employing advanced search functionality and auto-annotation capabilities through machine learning algorithms such as TotalSegmentator and Body Part Regression, the tool significantly reduces the manual labor required for image curation. 
Moreover, the tool's ability to support quality control and review mechanisms is vital for ensuring the reliability of datasets, especially in multi-center studies. 
The integration of a metadata dashboard is particularly noteworthy, as it enables the detection of potential biases in datasets. 
Furthermore, the open-source nature of the tool promotes collaboration and sharing among researchers, which is essential for advancing medical imaging research. 

By leveraging Kaapana's federated learning capabilities, in future work curated datasets can be used in downstream federated learning use cases, enabling a collaborative approach to machine learning that respects data privacy and locality constraints. 
While the use cases demonstrate the utility of the tool, quantifying its enhancements remains a primary focus for future work.
Furthermore, integrating even more advanced algorithms for automatic image annotation could further improve the efficiency and accuracy of the tool. 
Another potentially promising advancement could be the integration of Electronic Health Record (EHR) data, which plays a crucial role in the process of creating datasets.

These future directions aim to ensure that the Kaapana data curation tool remains at the forefront of medical imaging research, catering to the evolving needs of radiologists and machine learning researchers.


\section*{Acknowledgments}
Funded by "NUM 2.0" (FKZ: 01KX2121)

%
%
%
%

\bibliography{bib}
\end{document}